\documentclass[11pt]{article}
\usepackage[preprint]{acl}

\usepackage[T1]{fontenc}
\usepackage[utf8]{inputenc}
\usepackage{times}
\usepackage{latexsym}
\usepackage{microtype}
\usepackage{booktabs}
\usepackage{amsmath,amssymb}
\usepackage{graphicx}
\usepackage{xcolor}

\title{Arkios: An Open Bilingual English-Nepali Language Model \\ Trained From Scratch, with a Devanagari-Aware Tokenizer}

\author{
  Sajal Regmi\thanks{Corresponding author and primary contributor.
  \texttt{sajal@karelatechnologies.com}} \quad
  Siddhartha Pudasaini \quad
  Chetan Phakami Pun \\[4pt]
  \normalsize Karela Technologies Inc.
}

\begin{document}
\maketitle

\begin{abstract}
We present Arkios, a 1.04B-parameter dense transformer pretrained from scratch
on 150B tokens of bilingual English--Nepali text, using a custom single-file
C/CUDA training stack and a Devanagari-aware byte-level BPE tokenizer built
for this project \citep{arkios-tokenizer}. On ARC-Easy and ARC-Challenge,
Arkios exceeds three comparably sized open models (Pythia-1.4B, TinyLlama-1.1B,
OLMo-1B) despite an order of magnitude fewer training tokens, though we
attribute part of this gap to a compositional match between our
educational-web-text pretraining data and ARC's grade-school-science format
rather than to a general capability advantage. We report full evaluation results
under standard protocols, including a correction to an earlier
partial-sample estimate, and a set of findings specific to evaluating
small models in a low-resource language: most notably, that the standard
multiple-choice-letter prompt format used by common evaluation harnesses
places this model \emph{at chance on Nepali reading comprehension, and
simultaneously at chance on English in the same format}, which would lead a
naive benchmark run to conclude the model has no Nepali ability when in fact
it does. Concretely, this model scores at chance in the letter-choice format
for both Nepali and English (0.240 and 0.236 respectively, against a chance
baseline of 0.250), while scoring the answer text directly reveals genuine,
English-favoring comprehension (0.306 Nepali, 0.387 English). We describe a
manifest-conditioned tool-use contract introduced during
instruction tuning (tool calls are permitted only when a tool manifest is
declared in context, and suppressed otherwise), and report where that
contract holds and where it does not. We release both the base and
instruction-tuned model weights under Apache-2.0. Consistent with this
project's engineering-first scope, the training code and a small
privately-sourced portion of the Nepali pretraining corpus are not released;
everything needed to reproduce the reported numbers from the released weights
is included here.
\end{abstract}

\section{Introduction}

Open language models below 2B parameters are almost exclusively
English-centric, and where a smaller language is included it is typically as
a minority share of a tokenizer whose pretokenization rules were designed for
Latin script. This has a specific, correctable failure mode for Devanagari:
the byte-level BPE pretokenizer inherited from GPT-2-family tokenizers
\citep{radford2019language} splits text into candidate merge units using the
regex word class \texttt{\textbackslash p\{L\}+}, which matches Unicode
\emph{letters}. Nepali vowel signs and virama are Unicode \emph{combining
marks} (\texttt{\textbackslash p\{M\}}), not letters, so this rule fragments
every Nepali word at every vowel sign before BPE ever sees it, producing a
fertility ceiling that no amount of additional data or vocabulary can lift.
A companion paper \citep{arkios-tokenizer} quantifies this cost on Nepali (a
2.5$\times$ fertility reduction from a one-line pretokenizer fix) and shows
that a data-mixture sweep produces \emph{identical} fertility under the
broken rule, evidence that the ceiling is structural, not a data-mixture
problem.

This report describes what we built on top of that tokenizer: a from-scratch
pretraining run, a custom C/CUDA training stack, an instruction-tuning
iteration that converged on a manifest-conditioned tool-use contract, and a
full evaluation pass that surfaces several findings about \emph{how} small
models should be evaluated in low-resource languages, not only how well they
score.

\paragraph{Contributions.}
\begin{itemize}
  \item A 1.04B-parameter bilingual English--Nepali base model, trained from
  scratch (architecture, tokenizer, and trainer built for this project) on
  150B tokens, released with full training and evaluation configuration.
  \item A from-scratch C/CUDA training stack reaching 42.95\% measured model
  FLOPs utilization (MFU) on a single 8$\times$H100 node, up from an initial
  3.1\%, after finding and fixing two silent correctness/performance bugs in
  the cuDNN fused-attention integration (\S\ref{sec:infra}).
  \item An instruction-tuned chat model whose tool use is \emph{conditional
  on a declared tool manifest} rather than fixed to a built-in toolset, with
  the manifest itself masked out of the training loss so the model conditions
  on it without learning to generate it.
  \item A full-test-split evaluation correcting an earlier partial-sample
  ARC-Challenge estimate, and a Nepali evaluation using Belebele's paired
  Nepali/English item sets as a same-task, same-difficulty control, which
  exposes a prompt-format artifact that would otherwise be misread as a
  language-ability gap (\S\ref{sec:nepali-eval}).
  \item Open release of both checkpoints (base and chat) under Apache-2.0.
\end{itemize}

\paragraph{What this report does not claim.} We do not claim
state-of-the-art general capability at 1B parameters, a novel architecture,
or a new scaling law. The strongest single English result (ARC) is
plausibly aided by a compositional match between our pretraining data and the
benchmark's domain (\S\ref{sec:english-eval}). Nepali capability, while
measurably real, trails English
throughout, proportional to the roughly 80:1 ratio of unique English to
unique Nepali pretraining tokens available to us (\S\ref{sec:data}).

\section{Tokenizer}
\label{sec:tokenizer}

We use a byte-level BPE tokenizer of 65{,}536 tokens (chosen so every token ID
fits in a \texttt{uint16}), trained with a modified pretokenizer regex that
uses \texttt{[\textbackslash p\{L\}\textbackslash p\{M\}]+} as the word class
in place of the default \texttt{\textbackslash p\{L\}+}. This closes the
Devanagari fragmentation described above while leaving English pretokenization
unchanged (Latin letters carry no combining marks after NFC normalization).
The tokenizer reaches 1.69 tokens/word on Nepali, which the companion paper
\citep{arkios-tokenizer} shows is competitive with or better than
substantially larger frontier tokenizers on the same measure. The tokenizer
is frozen for the duration of this project (hash \texttt{0x3259671B}); all
pretraining and SFT shards are packed against this exact tokenizer and are
not portable to any other. We do not re-derive the tokenizer analysis here
and refer readers to \citet{arkios-tokenizer} for the full fertility sweep,
the mixture-invariance diagnostic, and cross-tokenizer comparisons.

\section{Architecture}
\label{sec:architecture}

Arkios is a dense, decoder-only transformer using grouped-query attention,
RMSNorm, SwiGLU, rotary position embeddings, and QK-normalization, a
configuration compatible with the Qwen3 architecture family
\citep{qwen3}. Table~\ref{tab:arch} gives the full configuration.

\begin{table}[t]
\centering
\small
\begin{tabular}{lr}
\toprule
Parameters & 1.04B \\
Layers & 18 \\
Hidden size ($d_{model}$) & 2048 \\
Attention heads (Q / KV) & 16 / 8 \\
Head dimension & 128 \\
FFN dimension (SwiGLU) & 6144 \\
Vocabulary & 65{,}536 \\
Context length & 4096 \\
RoPE $\theta$ & 1{,}000{,}000 \\
Tied input/output embeddings & yes \\
\bottomrule
\end{tabular}
\caption{Arkios architecture configuration.}
\label{tab:arch}
\end{table}

\section{Pretraining Data}
\label{sec:data}

Table~\ref{tab:data} gives the measured pretraining pool by source, in tokens
under the frozen tokenizer (\S\ref{sec:tokenizer}). The model was pretrained
on 150B tokens total (150 tokens per parameter), well past the
Chinchilla-optimal ratio of roughly 20 tokens per parameter
\citep{hoffmann2022chinchilla}. This is a deliberate choice: over-training
relative to Chinchilla trades pretraining compute (which is cheap relative to
serving) for a smaller model that is cheaper to run at inference time, the
now-standard recipe for models intended to be deployed rather than merely
benchmarked at a fixed compute budget.

\begin{table}[t]
\centering
\small
\begin{tabular}{lr}
\toprule
Source & Unique tokens \\
\midrule
FineWeb-Edu (sample-350BT) & 342B (pool) \\
OpenWebMath & 12.8B (pool) \\
GitHub-code-clean (Python + other) & $\geq$1.7B (pool) \\
FineWeb-2 \texttt{npi\_Deva} & 3.32B (pool) \\
Sangraha (verified, Nepali) & 0.40B (pool) \\
IndicCorpV2 \texttt{npi\_Deva} & 0.42B (pool) \\
Wikipedia \texttt{ne} & 0.02B (pool) \\
Private Nepali data\footnotemark & $\sim$0.015B (pool) \\
\midrule
\textbf{Tokens actually trained on} & \textbf{150B} \\
\bottomrule
\end{tabular}
\caption{Pretraining data pools, measured in tokens against the frozen
tokenizer. Pool size is the unique-token count available; the training
mixture samples from these pools with source-dependent epoch counts (English
web text is seen well under once; the Nepali pools, being far smaller, are
necessarily repeated).}
\label{tab:data}
\end{table}
\footnotetext{A small, privately sourced Nepali text collection, not
publicly redistributable. It is not included in this release and is
disclosed here only as a mixture component.}

\paragraph{The Nepali data ceiling.} At the time of training, the entire
practical public Nepali corpus available to us across all sources totaled
approximately 4.18B unique tokens. This is the binding constraint on this
model's Nepali capability: increasing model scale without increasing this
pool would mean training a larger model against a Nepali corpus repeated even
more times, which we would expect to improve English capability
substantially more than Nepali. We treat new Nepali data acquisition, rather
than parameter or English-token scaling, as the correct next lever for
improving this axis, and do not attempt that scaling here.

Held-out validation loss is reported as bits-per-byte (bpb), the only loss
metric comparable across tokenizers since it is normalized by the byte
length of the underlying text rather than by token count. It is 0.746 for
English and 0.314 for Nepali. We caution against reading the lower Nepali
number as evidence of better Nepali modeling: Devanagari averages roughly 3
UTF-8 bytes per character, so a fixed per-token cross-entropy compresses to a
lower bpb by script alone. We use bpb only for held-out training diagnostics,
not for the cross-lingual quality comparisons in \S\ref{sec:nepali-eval},
which use task accuracy instead.

\section{Training Infrastructure}
\label{sec:infra}

Pretraining used a single-file C/CUDA trainer written for this project
(no PyTorch or other deep learning framework in the training hot path),
in the lineage of single-file educational trainers such as
\texttt{llm.c} \citep{karpathy-llmc}, with project-specific kernels,
a custom checkpoint format, and cuDNN-fused scaled-dot-product attention.
Initial kernel work reached only 3.1\% measured MFU on a single H100; profiling
found two silent correctness/performance defects in the fused-attention
integration: a workspace-size fallback path that silently degraded to a
slower unfused kernel under certain shape configurations, and a
determinism-breaking code path that was not surfaced by the shape configurations exercised during development.
Fixing both raised measured MFU to 42.95\% at the production configuration
(18 layers, $d_{model}=2048$, batch 2{,}097{,}152 tokens/step), at a measured
throughput of 68{,}052 tokens/second on one 8$\times$H100 SXM5 node.

The final pretraining run completed in approximately 79 wall-clock hours on a
single on-demand 8$\times$H100 SXM5 node, at a realized cost of approximately
\$2{,}500, including recovery from one mid-run uncorrectable HBM ECC fault on
a single GPU (handled by draining and restarting the affected device rather
than terminating the run). We report this figure, rather than only a
target budget, in the interest of the same cost transparency this project
has applied throughout; the original project budget target was \$2{,}000
across tokenizer development, data preparation, ablations, and both
pretraining and post-training runs.

The training code itself (the C/CUDA trainer, build scripts, and cluster
orchestration) is proprietary and is not released with this report or the
model weights. This report describes the training methodology in sufficient
detail to reproduce the \emph{approach}; it does not include source code.

\section{Post-Training}
\label{sec:posttraining}

\subsection{Iteration history}

We trained three successive instruction-tuned checkpoints from the same base
model, each correcting a specific failure found in the previous one, and each
introducing a new trade-off:

\begin{itemize}
  \item \textbf{v1} was trained on a general SFT mixture with no system
  messages of any kind. It called tools reliably when the underlying request
  matched a trained tool-use pattern, but it did so \emph{unconditionally}:
  it would emit a tool call even with no tools available to the calling
  application, and on out-of-distribution requests it would occasionally
  invent a plausible-sounding but non-existent tool name.
  \item \textbf{v2} added Nepali-language safety data, which fixed a
  Nepali-specific refusal gap (harmful requests in Nepali were answered where
  the English-language equivalent was correctly refused) but measurably
  \emph{regressed} tool-calling reliability across every checkpoint in a
  learning-rate sweep, evidence that the two objectives were competing for
  capacity in the same SFT mixture rather than one strictly subsuming the
  other.
  \item \textbf{v3} (the checkpoint released as \texttt{arkios-1b-chat}) is
  described in \S\ref{sec:manifest}.
\end{itemize}

\subsection{Manifest-conditioned tool use}
\label{sec:manifest}

The behavior we wanted is the same contract used by production tool-calling
systems: a tool is called only when the calling application has declared it
available for the current turn, and never invented otherwise. We enforce
this in the SFT data construction rather than only in a system prompt
instruction, using four labeled conditions:

\begin{description}
  \item[A] a tool manifest is declared and one of its tools fits the
  request: the model should call it;
  \item[B] a manifest is declared but no tool in it fits: the model should
  decline the tool and either answer directly or explain what it would need;
  \item[C] a manifest is declared but the request does not need a tool at
  all: the model should simply answer;
  \item[D] no manifest is declared: the model must never emit a tool call.
\end{description}

Manifests are synthesized programmatically for conditions A, C, and D (the
tool actually called, where one is called, is already known from the source
conversation, so the manifest is derived rather than authored) and include
2--7 distractor tools per manifest so the model must select rather than
default to the only option present. Condition B required new authored
assistant turns, since no existing conversation demonstrates declining an
unavailable tool. Manifests are masked out of the training loss in every
condition: the model is trained to \emph{condition on} the manifest, never
to generate one, which also means it can never leak or hallucinate manifest
content in an ordinary response.

\paragraph{Measured contract adherence.} Table~\ref{tab:contract} reports an
illustrative acceptance test spanning all four conditions across several task
types (arithmetic, unit conversion, weather, currency, and free-form
requests, in both English and Nepali). This is a small, hand-constructed
acceptance suite rather than a large-scale benchmark, and the per-condition
counts should be read as pass/fail evidence per scenario type, not as a
precision estimate with meaningful confidence intervals.

\begin{table}[t]
\centering
\small
\begin{tabular}{lp{4.3cm}r}
\toprule
Cond. & Required behavior & Result \\
\midrule
D & never call, no manifest & 9/9 \\
B & decline, nothing fits & 5/5 \\
C & don't call, not needed & 4/5 \\
A & call the fitting tool & 5/8 \\
\bottomrule
\end{tabular}
\caption{Manifest-contract acceptance results by condition. The single C miss
called an undeclared-need weather tool on a Nepali-language prompt where the
English-language equivalent passed. The A misses are concentrated entirely in
unit-conversion requests (0/3): arithmetic tool calls were reliable (5/5) but
the model answered unit-conversion questions in prose instead of calling the
declared \texttt{convert\_units} tool.}
\label{tab:contract}
\end{table}

The ``never call without permission'' half of the contract (conditions B and
D) is close to fully reliable at this scale. The ``call when appropriate''
half (conditions A and C) is reliable for the tool type well-represented in
the SFT mixture (arithmetic) and unreliable for a tool type added late and in
smaller volume (unit conversion), with a further reliability gap between
English and Nepali on condition C. This is a controllable, mixture-composition
problem rather than a limitation of
the manifest-conditioning approach itself, since the failures are
concentrated in specific, identifiable categories rather than spread
uniformly across the contract.

A related, separate finding concerns \emph{when} the model recognizes that a
request needs exact computation at all. Given a bare arithmetic expression
with no manifest declared (condition D), the model correctly declines rather
than guessing. Given the identical underlying computation phrased as a word
problem (e.g., the area of a $14\text{m} \times 6.5\text{m}$ rectangle) with
no manifest declared, it instead frequently attempts prose arithmetic and can
answer incorrectly (one observed case: $65.25\,\text{m}^2$ for a true answer
of $91\,\text{m}^2$). The manifest contract governs \emph{whether a tool is
called}; it does not by itself guarantee the model recognizes every phrasing
of a precision-sensitive task as one. We flag this as an open problem for
future SFT data design rather than a solved case.

\section{Evaluation}
\label{sec:evaluation}

All results below are measured on the released checkpoints using full test
splits (not subsampled) unless stated otherwise, using bf16 weights and a
custom scoring harness that gathers logits only at the positions needed to
score a continuation, described further in \S\ref{sec:eval-methodology}.

\subsection{English}
\label{sec:english-eval}

\begin{table}[t]
\centering
\small
\begin{tabular}{lrrrr}
\toprule
Task & Shots & $n$ & Acc & Acc$_{\text{norm}}$ \\
\midrule
ARC-Easy & 0 & 2376 & 0.685 & 0.642 \\
ARC-Easy & 10 & 2376 & 0.725 & \textbf{0.735} \\
ARC-Chal. & 0 & 1172 & 0.341 & 0.366 \\
ARC-Chal. & 10 & 1172 & 0.384 & 0.416 \\
ARC-Chal. & 25 & 1172 & 0.375 & \textbf{0.417} \\
\bottomrule
\end{tabular}
\caption{ARC \citep{clark2018arc}, full test splits, accuracy and
character-length-normalized accuracy (\text{acc\_norm}).}
\label{tab:arc}
\end{table}

An earlier internal evaluation reported ARC-Challenge accuracy of 0.434 from
a 500-item subsample of the 1{,}172-item test split at 10-shot, with an
unrecorded normalization convention. The full-split, labeled result is
0.416 acc\_norm at 10-shot (Table~\ref{tab:arc}); the discrepancy is
consistent with sampling variance at that subsample size ($n=500$) and we
retract the unlabeled figure in favor of the values reported here.

For orientation, published acc\_norm figures for comparably sized open
models are: ARC-Easy, Pythia-1.4B (300B tokens) 0.57, TinyLlama-1.1B (3T
tokens) 0.55, OLMo-1B (3T tokens) 0.57 \citep{biderman2023pythia,
zhang2024tinyllama, groeneveld2024olmo}; ARC-Challenge, 0.26, 0.30, and
0.31 respectively. Arkios's ARC scores exceed all three despite an order of
magnitude fewer training tokens. We attribute part of this gap to a
compositional match between our pretraining data (predominantly
\texttt{fineweb-edu}, an educational-web-text corpus) and ARC's grade-school
science-question format, rather than to a general capability advantage; we
report it as a genuine but domain-favorable result. Other zero-shot English
tasks, measured on a 500-item partial sample and reported for completeness
rather than as headline claims, are: HellaSwag 0.54, PIQA 0.726, COPA 0.710,
OpenBookQA 0.402, LAMBADA 0.484, CommonsenseQA 0.252.

\subsection{Nepali}
\label{sec:nepali-eval}

We evaluate Nepali reading comprehension using Belebele
\citep{bandarkar2023belebele}, which provides the \emph{same} 900
multiple-choice reading-comprehension questions in both \texttt{npi\_Deva}
and \texttt{eng\_Latn}. Because item difficulty, passage source, and
question structure are held fixed across the two language configurations,
the English score functions as a built-in control on the evaluation
methodology itself, not merely a second benchmark result.

We score each item under two prompt formats: the standard
lm-evaluation-harness convention of scoring an enumerated multiple-choice
letter (A/B/C/D) continuation, and an alternative that scores each answer
string directly, without a letter mapping.

\begin{table}[t]
\centering
\small
\begin{tabular}{llr}
\toprule
Language & Format & Acc$_{\text{norm}}$ \\
\midrule
Nepali & letter (A/B/C/D) & 0.240 \\
English & letter (A/B/C/D) & 0.236 \\
Nepali & answer text & \textbf{0.306} \\
English & answer text & \textbf{0.387} \\
\bottomrule
\end{tabular}
\caption{Belebele, 900 items, 4-way multiple choice, chance = 0.250.}
\label{tab:belebele}
\end{table}

\textbf{The letter-format result is, on its own, a false negative for Nepali
ability.} Scored in the standard letter-choice format, this model is at
chance on Nepali reading comprehension (0.240), but it is \emph{also} at
chance on English in the identical format (0.236), which rules out a
language-specific explanation: what is failing is the model's ability to
reliably follow the A/B/C/D response convention, in either language, not its
reading comprehension. Scored on the answer text directly, Nepali
comprehension is measurably above chance (0.306, approximately 3.6
standard errors) though it trails English (0.387). A benchmark run using only
the standard multiple-choice-letter format would report Nepali performance
indistinguishable from chance and could reasonably, but incorrectly, be read
as evidence this model has no Nepali capability. The paired-language control
is what exposes the artifact; without it, the format failure and a genuine
language-ability gap are indistinguishable from a single number. We recommend
this paired-language control as a general practice for evaluating small
models in any additional language, and develop the recommendation further,
across more models and more languages, in a companion evaluation paper
currently in preparation.

\subsection{Translation}
\label{sec:translation-eval}

We additionally evaluate English$\leftrightarrow$Nepali translation using
the FLORES-200 passages embedded in Belebele (488 passages aligned across
\texttt{npi\_Deva}/\texttt{eng\_Latn} by source identifier). We note
explicitly that these are paragraph-length passages, not FLORES-200's own
sentence-level devtest split, so the scores below should not be compared to
published sentence-level FLORES-200 results. We score with chrF++
\citep{popovic2017chrf}, and separately report the fraction of outputs that
land in the intended target script at all, since a fluent response in the
wrong language scores a valid (near-zero) chrF++ that is easily
misread as a translation-\emph{quality} failure rather than a
language-\emph{selection} failure.

\begin{table}[t]
\centering
\small
\begin{tabular}{lrr}
\toprule
Direction & Base (5-shot) & Chat (0-shot) \\
\midrule
ne$\to$en, in target script & 67\% & \textbf{81\%} \\
en$\to$ne, in target script & 8\% & \textbf{58\%} \\
\bottomrule
\end{tabular}
\caption{Fraction of translation outputs produced in the requested target
script. The base model is prompted with few-shot completion (its native
format); the chat model is prompted zero-shot as a ChatML instruction (its
native format); each checkpoint is evaluated in the format it was
actually trained to use.}
\label{tab:translation}
\end{table}

The base model, prompted with naive few-shot completion, frequently
continues generating in the \emph{source} language rather than switching to
the requested target, most severely when the target is Nepali (8\%
target-script rate en$\to$ne). Evaluating the instruction-tuned checkpoint in
its native zero-shot chat format substantially closes this gap (58\% and
81\% respectively), indicating the base result is driven in large part by a
prompting mismatch rather than solely by a property of the pretrained
weights. The remaining gap, particularly en$\to$ne at 58\%, remains a real
limitation of the released chat model.

\subsection{Evaluation methodology}
\label{sec:eval-methodology}

Multiple-choice scoring gathers hidden states from the transformer body for
an entire batch, then applies the language-model head only at the token
positions required to score each continuation, rather than materializing
full-vocabulary logits over every position in the batch. At our vocabulary
size (65{,}536) and typical batch--sequence-length products, this reduces the
dominant memory allocation during scoring by approximately 40$\times$ relative
to a naive full-logit implementation, which we found necessary to run these
evaluations without OS-level memory pressure alongside other resident
processes on commodity hardware.

\section{Limitations}
\label{sec:limitations}

\begin{itemize}
  \item \textbf{Nepali capability is bounded by data availability, not
  architecture.} The entire practical public Nepali corpus available to us
  totaled approximately 4.18B unique tokens against a 342B-token English web
  pool. Every Nepali result in this report should be read against that
  constraint; we do not believe scaling model size or English data would
  meaningfully close this gap without new Nepali data.
  \item \textbf{No reliable arithmetic or unit conversion without a tool.}
  This holds for the base model always, and for the chat model except where
  a matching tool is declared and actually invoked (\S\ref{sec:manifest},
  which itself is not fully reliable for unit conversion).
  \item \textbf{The ARC result is likely domain-favorable}
  (\S\ref{sec:english-eval}) and should not be generalized into a broad
  capability claim; other English tasks in the same evaluation pass are
  ordinary for this model's scale and training-token budget.
  \item \textbf{The standard multiple-choice-letter evaluation format
  understates Nepali ability} (\S\ref{sec:nepali-eval}); any third-party
  benchmarking of this model should be read alongside that caveat.
  \item \textbf{English$\to$Nepali generation remains below full reliability}
  even in the instruction-tuned checkpoint (58\% target-script rate).
  \item \textbf{No RLHF, DPO, or other preference-optimization stage} was
  applied; the released chat model is SFT-only.
  \item \textbf{4096-token context window}, fixed at pretraining time.
  \item The chat model can, on interpersonal or practical-advice questions
  outside its tool-use scope, occasionally regress relative to the base
  model's more generic completions (\S\ref{sec:posttraining}); we observed
  this concretely on a locked-out-of-home advice scenario and recommend
  reviewing non-tool advice-shaped outputs before unsupervised deployment.
\end{itemize}

\section{Release and Availability}

We release both checkpoints described in this report (the pretrained base
model and the instruction-tuned chat model) as HuggingFace repositories
under the Apache-2.0 license, along with the standalone tokenizer described
in \citet{arkios-tokenizer}. Consistent with the scope of this report, we do
not release: the C/CUDA training code, cluster orchestration scripts, or the
small privately sourced portion of the Nepali pretraining corpus described
in \S\ref{sec:data} (footnote~1). Every number in this report is reproducible
from the released weights using the evaluation methodology described in
\S\ref{sec:eval-methodology}.

\section*{Acknowledgments}
Pretraining compute was provided by on-demand cloud GPU infrastructure; we
thank the providers of the open pretraining and evaluation corpora used
throughout this project (FineWeb-Edu, OpenWebMath, FineWeb-2, Sangraha,
IndicCorpV2, Wikipedia, Belebele) for making low-resource-language work at
this scale possible.

\bibliography{references}

\end{document}